\documentclass[letterpaper, 10 pt, conference]{ieeeconf}  

\IEEEoverridecommandlockouts                              
\usepackage{algorithm}
\usepackage{algpseudocode}
\usepackage{amsmath}
\usepackage{times}
\usepackage{amssymb}
\usepackage{booktabs}
\usepackage{physics}
\usepackage{graphicx}
\usepackage{url}
\usepackage[table]{xcolor}
\usepackage{float}
\usepackage{dblfloatfix}
\usepackage{subfig}
\usepackage{textcomp}
\usepackage{gensymb}
\usepackage[normalem]{ulem}
\usepackage{multicol,tabularx,capt-of}
\usepackage{multirow}
\usepackage[hidelinks]{hyperref}
\usepackage[absolute,overlay]{textpos}
\usepackage{booktabs}
\usepackage{tabularx}
\usepackage{siunitx}
\usepackage{threeparttable}
\newcommand{\valpm}[2]{%
  \ensuremath{\scalebox{0.95}{\ensuremath{#1}}%
             \,{\scalebox{0.85}{$\pm$}}\,%
             \scalebox{0.85}{\ensuremath{#2}}}%
}

\begin{document}
\title{\LARGE \bf
PMTRM: Pseudo-Memory Temporal Re-encoding Module for Embodied Policy Learning
}

\author{Changchuan Yang$^{1}$, Haoxuan Xu$^{2}$, Wenbo Chen$^{3}$, Shuai Ren$^{4}$,\\ Jianlong Zheng$^{5}$, Huarui Zhang$^{3}$, Tianfu Li$^{3}$, Guanzhong Tian$^{1,6}$
\thanks{*This work is supported in part by Ningbo Science and Technology Innovation Yongjiang 2035 Key Technology  Project, 2024Z265, and in part by Ningbo Key Research and Development Plan under Grant 2025Z105.}%
\thanks{$^{1}$Zhejiang University}%
\thanks{$^{2}$The Hong Kong University of Science and Technology}%
\thanks{$^{3}$The Hong Kong University of Science and Technology (Guangzhou)}%
\thanks{$^{4}$vivo Robotics Lab}%
\thanks{$^{5}$South China Normal University}%
\thanks{$^{6}$Corresponding author: {\tt\small gztian@zju.edu.cn}}%
}

\makeatletter

\makeatother
\maketitle
\thispagestyle{empty}
\pagestyle{empty}

\begin{abstract}
Robotic manipulation often contains repeated motions whose local observations look similar at different phases.
When these phases require different actions, a policy that relies mainly on the current observation may repeat completed motions or switch phases at the wrong time. To address this phase ambiguity, we present the Pseudo-Memory Temporal Re-encoding Module (PMTRM), a lightweight plug-in module with only 7.61M parameters that encodes a bounded history of executed states and actions into a latent sequence for existing policies. To help distinguish phases, a temporal heterogeneity objective penalizes positive similarity between distant positions in this sequence, while anchor and reconstruction losses preserve information needed for action prediction. The reconstruction decoder is used only during training, leaving the temporal re-encoder to supply history to the policy at inference. We train the module progressively on synthetic sequences and robot data, then jointly with the policy, using temporal masking to accommodate partial histories. This integration retains the original action head and action space and adds auxiliary losses to the original policy loss. Experiments with multiple policy backbones in simulation and on a real robot show improved task success on tasks with phase ambiguity, with little additional computation.
\end{abstract}


\section{INTRODUCTION}
In robotic manipulation, the same or similar observation may occur at different execution phases that require different actions.
For example, a button may look unchanged after a press~\cite{shi2025memoryvla}, and a robot moving along the same path in opposite directions may need history to identify the current phase~\cite{hu2025pam}.
We use \emph{phase ambiguity} to describe this situation: observations from different execution phases are locally similar, but the appropriate next action depends on the preceding execution history.
A policy without sufficient history may therefore repeat a completed motion or switch phases at the wrong time.
Recent work on cyclic manipulation further shows the need to track progress when an action must be repeated a specified number of times~\cite{wei2026cyclemanip}.
Reliable action selection in these settings therefore requires a representation of execution history that distinguishes phases with similar observations.

\begin{figure}[t]
\centering
\includegraphics[width=0.5\textwidth]{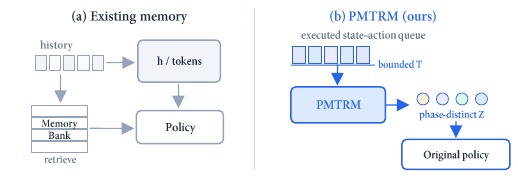}
\caption{
Conceptual comparison of temporal memory mechanisms for repetitive manipulation.
(a) Representative memory-based approaches summarize history into recurrent states or memory tokens, or retrieve relevant information from an explicit memory bank.
(b) PMTRM instead re-encodes a bounded queue of executed state--action history into a phase-discriminative latent sequence $Z$.
}
\label{fig:memory_comparison}
\vspace{-4mm}
\end{figure}

Recent imitation learning and embodied policy frameworks have substantially improved robotic action prediction, with representative examples including ACT~\cite{zhao2023learning}, Diffusion Policy~\cite{chi2023diffusion}, and vision-language-action models such as OpenVLA~\cite{kim2024openvla}. Nevertheless, resolving phase ambiguity requires temporal representations that preserve relevant execution history and distinguish phases with similar observations. Existing memory mechanisms commonly provide this context through recurrent states~\cite{gui2026seedpolicy}, learned history tokens~\cite{hu2025pam}, or retrieval from a memory bank~\cite{shi2025memoryvla}, as conceptually shown in Fig.~\ref{fig:memory_comparison}(a). These mechanisms support decisions that depend on past interactions, but maintaining and accessing memory can require dedicated update rules, retrieval components, or policy-specific fusion layers. Such requirements increase integration effort across policy backbones, even when the relevant phase information is contained in a bounded recent history. The challenge is therefore to preserve action information and distinguish execution phases through a lightweight history representation that requires minimal changes to the policy. In practical robotic deployment, such a module should introduce limited computational overhead and remain scalable across different policy backbones rather than relying on additional memory banks or large architectural modifications.

To learn such a representation, we introduce PMTRM, which combines a temporal re-encoder, representation learning objectives, and a progressive training procedure.
The re-encoder maps a bounded history of executed states and actions to a latent sequence for the policy, as shown in Fig.~\ref{fig:memory_comparison}(b).
To discourage similar representations at distant time steps, we apply a temporal heterogeneity objective directly to this sequence.
Anchor and reconstruction losses preserve information needed for action prediction, balancing temporal discrimination with reconstruction of the input history.
These objectives use temporal distance rather than phase labels during training.
We first pretrain PMTRM on synthetic sequences, adapt it to robot data, and apply mirror augmentation and temporal masking before joint policy training.
Joint training adds the auxiliary objectives to the original policy loss.
At inference, only the re-encoder is retained, providing recent history through the policy input adapter while preserving the original action head and action space.
The main contributions of this work are summarized as follows:
\begin{itemize}
\item We propose PMTRM, a lightweight bounded-history temporal representation module with 7.61M parameters, whose policy-facing latent is explicitly regularized for distant-phase discriminability while remaining reconstructable.

\item We combine temporal heterogeneity with anchor and reconstruction losses to encourage temporal discrimination while preserving action information, without phase labels.

\item We evaluate PMTRM across action-policy backbones and phase-ambiguous tasks in simulation and on a real robot, and analyze the effects of the proposed objectives and mirror-mask training.

\end{itemize}

\section{Related Work}

\subsection{Temporal Memory and Pseudo-Memory}
History-aware policies encode past interactions through recurrent states~\cite{gui2026seedpolicy}, working-memory tokens~\cite{hu2025pam}, or retrieved perceptual and cognitive features~\cite{shi2025memoryvla}. These mechanisms provide historical context, while cyclic manipulation also requires tracking execution progress~\cite{wei2026cyclemanip}. PMTRM focuses on a bounded state-action history and directly regularizes its latent sequence to distinguish distant temporal positions.



\subsection{Policy-Side Temporal Conditioning}
ACT predicts action chunks with a Transformer \cite{zhao2023learning}, while Diffusion Policy generates multimodal actions by denoising \cite{chi2023diffusion}. PMTRM neither replaces these policy mechanisms nor compresses history into a single summary: its two-stage design exposes the full bounded latent sequence to the unchanged action head and uses a separate decoder only during training. It is preferable when recent phase aliasing matters and a low-latency, backbone-agnostic adapter is needed; explicit memory remains preferable for semantic retrieval or context beyond the queue.


\subsection{Temporal Discriminability}

Prior work learns temporally structured representations through temporal correspondence, sequence alignment, or future prediction~\cite{sermanet2018time}, while robot-learning methods further exploit temporal organization for hierarchical actions and progress modeling~\cite{gao2025must,myers2025temporal}. PMTRM instead directly regularizes the bounded history representation consumed by the policy, suppressing similarity between temporally distant positions while preserving action-relevant information through anchor and reconstruction objectives. This design motivates the temporal heterogeneity objective introduced next.

\section{Method}

\begin{figure*}
    \centering
    \includegraphics[width=1.0\textwidth]{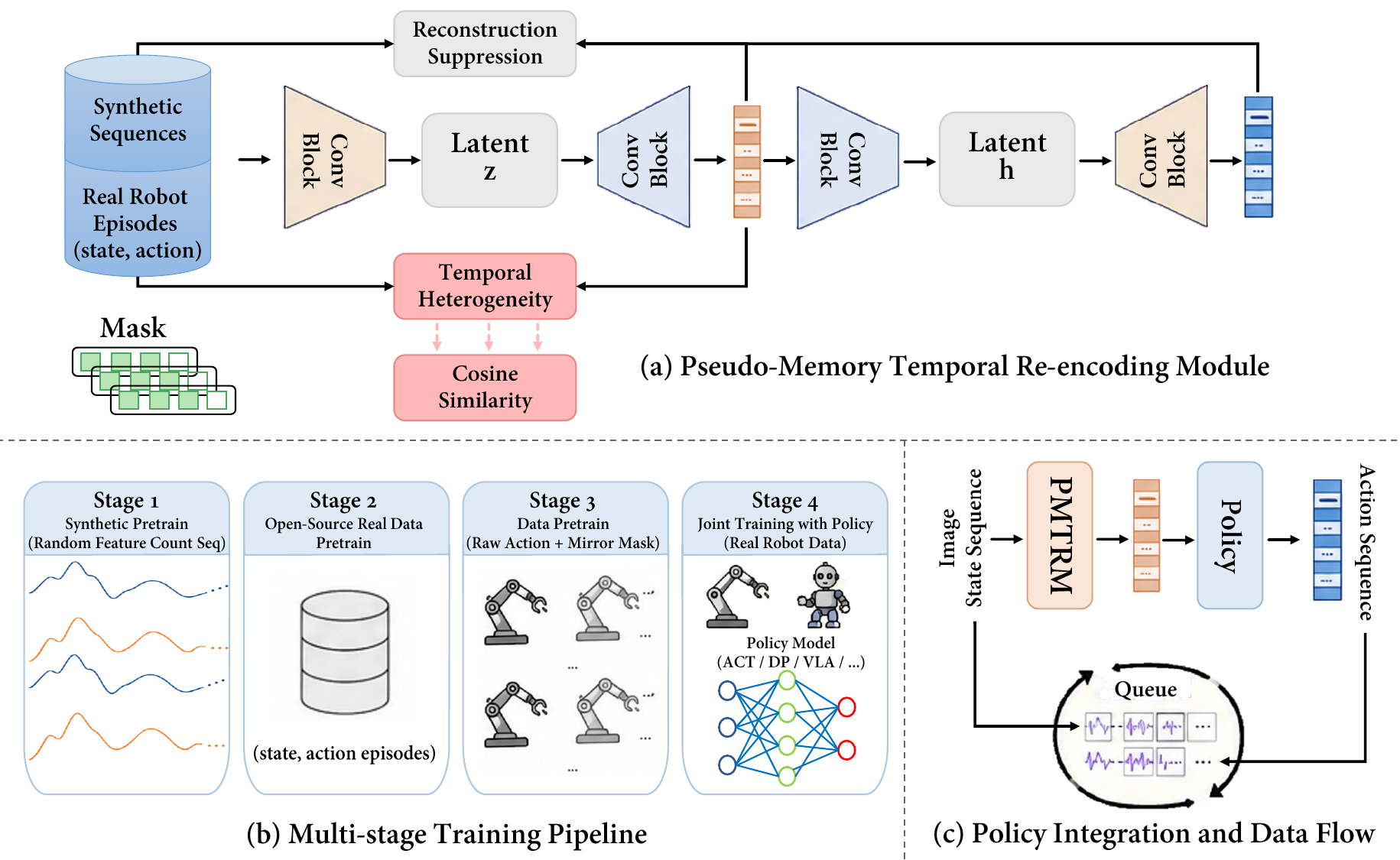}
    \vspace{-6mm}
    \caption{Overall framework of PMTRM.
(a) The module employs a two-stage convolutional temporal re-encoding autoencoder to map synthetic or real executed state-action histories into pseudo-memory representations.
(b) Stages 1--4 correspond to synthetic pretraining, open-source real-data fine-tuning, mirror-mask fine-tuning (Eq.~\eqref{eq:mirror}), and joint policy training, respectively.
(c) At deployment, $E_2$ is disabled and only $E_1$ encodes the executed state/action history queue (up to $T$ steps). The policy consumes this queue representation and predicts a separate future action chunk, which is not re-encoded until its actions are executed.}
    \label{fig:framework}
    \vspace{-4mm}
\end{figure*}

As illustrated in Fig.~\ref{fig:framework}, PMTRM has three components: a temporal re-encoder and its policy interface, temporal heterogeneity and information preservation objectives, and a progressive training procedure. We first define the history representation, then describe the objectives that shape it, and finally explain training and integration with downstream policies.

\subsection{Pseudo-Memory Temporal Re-encoding Module}

Given a state-action trajectory segment, we denote the input temporal sequence as
\begin{equation}
    \mathbf{X} = [\mathbf{x}_1, \mathbf{x}_2, \ldots, \mathbf{x}_T] \in \mathbb{R}^{T \times D},
    \label{eq:input}
\end{equation}
where $T$ is the sequence length and $D$ is the maximum feature dimension after padding. In practice, different episodes may have different valid lengths and feature dimensions.
Therefore, we use a temporal mask $\mathbf{m}^{t} \in \{0,1\}^{T}$ and a feature mask
$\mathbf{m}^{f} \in \{0,1\}^{D}$ to indicate valid time steps and valid feature channels.
The feature mask is episode-constant and records the available proprioceptive/action dimensions; only the temporal mask varies with the history prefix. For ACT and HACT-VQ, PMTRM receives proprioceptive state plus executed actions (no image tokens). For visual backbones, the original visual encoder produces one feature $\mathbf{v}_t$ for each queued frame; these features form $\mathbf{V}\in\mathbb{R}^{T\times D_v}$ and are temporally aligned with the same queue entries as $\mathbf{X}$. The full mask and re-encoding process are
\begin{equation}
\begin{aligned}
\mathbf{M}_{t,d} &= \mathbf{m}^{t}_{t} \mathbf{m}^{f}_{d}, \\
\mathbf{Z} &= E_1(\mathbf{X}, \mathbf{M}), \qquad
\hat{\mathbf{X}} = E_2(\mathbf{Z}, \mathbf{M}), \\
\mathbf{C} &= [\mathbf{Z}\,\Vert\,P_v(\mathbf{V})].
\end{aligned}
\label{eq:encoding}
\end{equation}

Here $P_v:\mathbb{R}^{D_v}\rightarrow\mathbb{R}^{D}$ is a learned linear projection and $\Vert$ denotes channel concatenation. DP and VLA policies consume the full, time-aligned $\mathbf{C}\in\mathbb{R}^{T\times 2D}$ through a backbone-specific input adapter, not only $\mathbf{z}_T$; non-visual policies use $\mathbf{C}=\mathbf{Z}$. Temporal masking is applied to PMTRM inputs and losses at every stage; the visual backbone receives its native frame-validity mask, and masked visual positions are zeroed after $P_v$ before concatenation. $E_1$ is policy-facing, while $E_2$ maps $\mathbf{Z}$ back to the trajectory space only during training.

Each $E_1/E_2$ stack has $L=4$ residual blocks with $k=3$, hidden width $H=128$, dilations $\{1,2,4,8\}$, GroupNorm, and GELU; dropout is disabled. Feature masks are active throughout pretraining and joint training; random prefix masking is used in all stages, whereas mirror augmentation is used only in the third fine-tuning stage.

Unlike a bottleneck, $\mathbf{Z}\in\mathbb{R}^{T\times D}$ preserves the input sequence shape; PMTRM is therefore a lightweight bounded-history representation, not a coordinate-compression method. Relative to a standard autoencoder, it exposes $\mathbf{Z}$ to the policy and heterogeneity loss while using $E_2$ only to preserve action-relevant information.


\subsection{Masked Reconstruction and Latent Anchor Loss}

The re-encoding module is trained under two information-preservation constraints. We call the first term the latent anchor loss. It defines a small residual trust region around the normalized input trajectory; it does not force an identity mapping, but prevents the heterogeneity objective from destroying action semantics:
\begin{equation}
    \mathcal{L}_{\mathrm{anchor}}
    =
    \mathrm{MSE}_{\mathbf{M}}(\mathbf{X}, \mathbf{Z})
    +
    0.1 \cdot
    \mathcal{L}_{\mathrm{cos}}(\mathbf{X}, \mathbf{Z}),
    \label{eq:anchor}
\end{equation}
where $\mathrm{MSE}_{\mathbf{M}}(\cdot)$ denotes masked mean squared error and $\mathcal{L}_{\mathrm{cos}}$ is a masked cosine distance computed over valid positions.
The weight of this term is intentionally small in the total objective. Thus, $\mathcal{L}_{\mathrm{anchor}}$ does not dominate representation learning; it prevents the re-encoder from producing arbitrary latent values that satisfy temporal heterogeneity but no longer correspond to executable state-action transitions.

The reconstruction term in Eq.~\eqref{eq:recon} is
\begin{equation}
\mathcal{L}_{\mathrm{recon}}=\mathrm{MSE}_{\mathbf{M}}(\mathbf{X},\hat{\mathbf{X}})+0.1\mathcal{L}_{\mathrm{cos}}(\mathbf{X},\hat{\mathbf{X}}),
\label{eq:recon}
\end{equation}
which preserves information needed for executable action prediction.

\subsection{Temporal Heterogeneity Objective}

To improve temporal discriminability in the re-encoded latent sequence, we introduce an intra-trajectory heterogeneity objective on the output of the first autoencoder. The motivation is that distant time steps within a trajectory should not collapse into overly similar latent representations, especially when the original action sequence contains repeated or locally similar motion patterns.

Given the latent sequence $\mathbf{Z} \in \mathbb{R}^{T \times D}$, we define the temporal heterogeneity objective as
\begin{equation}
\begin{aligned}
\tilde{\mathbf{z}}_t
&=
\frac{
\mathbf{z}_t \odot \mathbf{m}^{f}
}{
\sqrt{
\left\|
\mathbf{z}_t \odot \mathbf{m}^{f}
\right\|_2^2
+\epsilon
}}, \\
q_{k,t}
&=
\mathbf{m}^{t}_{t}
\cdot
\mathbb{I}
\left(
|t-a_k|>\delta
\right), \\
s_{k,t}
&=
\tilde{\mathbf{z}}_{a_k}^{\top}
\tilde{\mathbf{z}}_t, \\
\mathcal{L}_{\mathrm{hetero}}
&=
\frac{
\sum_{k=1}^{K}
\sum_{t=1}^{T}
q_{k,t}
\left[
\max(0,s_{k,t})
\right]^2
}{
\sum_{k=1}^{K}
\sum_{t=1}^{T}
q_{k,t}
+\epsilon
}.
\end{aligned}
\label{eq:hetero}
\end{equation}
Here, $\mathbf{m}^{f}$ is the feature mask, $\mathbf{m}^{t}$ is the temporal mask, $\epsilon$ is a small constant for numerical stability, and $\mathcal{A}=\{a_1,\ldots,a_K\}$ denotes randomly sampled valid anchor indices. The indicator $q_{k,t}$ excludes invalid temporal positions and local pairs within the margin $\delta$.

This objective penalizes positive similarity only for distant pairs; local smoothness is unconstrained while anchor and reconstruction protect action semantics. We use $K=32$ and $\delta=8$; anchors are resampled per batch. The $K\in\{16,32,64\}$ and $\delta\in\{4,8,12\}$ sweep was selected on synthetic pretraining and then checked on ACT Button Press-and-Release and visual DP Stack Block. It preserves the ordering and changes final success by at most 1.4 points, so the same setting is fixed for all backbones.
For short or heavily masked windows, anchors are sampled only from valid positions and the heterogeneity term is skipped when fewer than two distant valid pairs remain. Across training, 93.4\% of synthetic windows, 89.7\% of real-data windows, and 86.1\% of mirror crops contain at least eight active anchor--time pairs; the remaining batches contribute only anchor and reconstruction gradients. This prevents unstable large gradients when the available history is too short.
\subsection{Overall Training Objective}

The final training loss combines
temporal heterogeneity, latent anchoring, and reconstruction:
\begin{equation}
    \mathcal{L}
    =
    \alpha_{\mathrm{h}} \mathcal{L}_{\mathrm{hetero}}
    +
    \beta \mathcal{L}_{\mathrm{anchor}}
    +
    \gamma \mathcal{L}_{\mathrm{recon}}.
    \label{eq:loss}
\end{equation}
In our implementation, $\alpha_{\mathrm{h}}=1.5$, $\beta=0.05$, and $\gamma=8.0$. After normalization, typical batch values are $\mathcal{L}_{\mathrm{hetero}}=0.10$--0.16, $\mathcal{L}_{\mathrm{anchor}}=0.03$--0.06, and $\mathcal{L}_{\mathrm{recon}}=0.006$--0.012; thus $\gamma$ maintains reconstruction while $\beta$ bounds the residual.

\subsection{Progressive Pretraining and Fine-tuning}
\label{subsec:PPF}

The four stages in Fig.~\ref{fig:framework} are synthetic pretraining, real-data fine-tuning, mirror-mask fine-tuning, and joint policy training. The first three train PMTRM; the fourth attaches it to a backbone. Synthetic windows mix periodic and non-periodic signals with random lengths and active dimensions, padded and masked to the model shape.

Second, we fine-tune on LeRobot-format raw-action trajectories. Channels are ordered as pose, gripper, then action, normalized per dataset, and padded/masked to the maximum dimension; this canonical ordering avoids high-to-low-dimensional transfer degradation. We sample shuffled episode windows and cap large datasets per epoch so no corpus dominates gradients.

Synthetic pretraining uses 50{,}000 windows for 100 epochs (batch size 64, AdamW, $10^{-3}$). Length is uniform on $[32,128]$ and active dimensions on $[4,D]$: sinusoidal channels have amplitude $[0.2,1.0]$, 1--4 cycles/window, random phase, and Gaussian noise $\sigma\in[0,0.02]$; piecewise-smooth channels linearly interpolate four random knots with the same amplitude/noise ranges. Windows are channel-normalized. Real-data fine-tuning uses the same optimizer and batch size for 30 epochs at $3\times10^{-4}$, with at most 2{,}048 windows per dataset per epoch. Skipping synthetic pretraining lowers cyclic Button Press-and-Release CSR by 3.2 points and the cross-embodiment mean completion by 2.0 points.

Third, we use mirror-and-mask training as representation augmentation, not as a physically executable reverse rollout. Given an episode sequence $\mathbf{X}$, we construct the virtual stream
\begin{equation}
\mathbf{X}^{\mathrm{mir}}
    =
    [\mathbf{X}, \mathrm{rev}(\mathbf{X}), \mathrm{rev}(\mathbf{X}), \mathbf{X}].
    \label{eq:mirror}
\end{equation}
and sample a contiguous window of at most $T=128$ entries (then left-pad and mask it); the fourfold stream is never passed as a whole. For a three-step sequence $[A,B,C]$, the virtual order is $[A,B,C,C,B,A,C,B,A,A,B,C]$, so sampled crops can cross reversal boundaries. Reversal is applied to normalized vectors without sign-flipping channels because a universal physical inverse is undefined. The stream is never executed or a policy target; it only regularizes changed temporal order. Prefix length is uniform on $[1,\sum_t m^t_t]$.
Fourth, during downstream policy training, PMTRM is optimized jointly with the original policy objective using the same auxiliary loss. This stage adapts the latent representation to the action distribution and observation encoder of the selected policy backbone. 
An essentiality check on the same validation split gives average memory-task success of 67.7\% without Stage~1, 69.1\% without Stage~2, and 68.4\% when Stage~3 mirror augmentation is removed; the full four-stage schedule reaches 70.9\%. Thus synthetic diversity and mirror robustness are complementary, while Stage~2 mainly improves cross-embodiment calibration. A deployment requiring minimal training can omit Stage~3 at a cost of 2.5 points, but keeps the same inference graph.

\begin{table}[t]
\centering
\small
\caption{Open-source robot manipulation datasets used in the fine-tuning setting. The corpus includes single-arm and bimanual data (UMI and BEHAVIOR-1K), deformable-object interactions (towel/dish washing and soft-fold), and multi-stage trajectories (LIBERO, CALVIN, BEHAVIOR-1K, and X-VLA); these tags describe pretraining diversity rather than a separate benchmark split.}
\label{tab:additional_datasets}
\resizebox{\columnwidth}{!}{
\begin{tabular}{lrrr}
\toprule
Dataset family & State/Action & Episodes & Frames \\
\midrule
DROID/Bridge/RT-1~\cite{khazatsky2024droid,walke2023bridgedata,brohan2022rt1} & 7--8/7 & 137{,}700 & 5.6M \\
LIBERO/CALVIN~\cite{liu2023libero,mees2022calvin} & 8--15/7 & 34{,}800 & 2.6M \\
BEHAVIOR-1K~\cite{li2024behavior1k} & 256/23 & 10{,}000 & 119.1M \\
TACO/Franka/TOTO~\cite{rosetebeas2023tacorl,cui2023play,zhou2023toto} & 7--13/7--15 & 4{,}200 & 0.6M \\
Austin/UT/Berkeley/Stanford~\cite{liu2023sirius,nasiriany2023sailor,shah2023mutex,radosavovic2023rpt,belkhale2023hydra} & 7--15/7--8 & 3{,}800 & 1.8M \\
X-VLA soft-fold~\cite{zheng2026xvla} & 96/14 & 1{,}542 & 2.9M \\
UMI/RoboCOIN (bimanual/deformable)~\cite{chi2024umi,wu2025robocoin} & 7--26/7--26 & 1{,}755 & 1.1M \\

\bottomrule
\end{tabular}
}
\end{table}

\subsection{Integration with Downstream Embodied Policies}

PMTRM is a plug-in representation component for ACT, Diffusion Policy, and VLA policies: preceding executed states and actions are padded, encoded, and supplied to the original policy. It leaves continuous actions, discrete tokens, action chunks, decoders, and normalization unchanged.

When PMTRM is integrated with a downstream policy backbone, its auxiliary objective is directly added to the original policy training loss. 
The joint optimization objective is defined as
\begin{equation}
    \mathcal{L}_{\mathrm{joint}}
    =
    \mathcal{L}_{\mathrm{policy}}
    +
    \mathcal{L}_{\mathrm{PMTRM}},
    \label{eq:joint}
\end{equation}
where $\mathcal{L}_{\mathrm{policy}}$ denotes the original loss function of the corresponding policy model
All policy-specific objectives, schedules, and architectural settings follow each backbone's default configuration; PMTRM is jointly optimized without task-specific loss redesign.

\subsection{Computational Cost and Scaling}
PMTRM has linear cost $O(TLkDH)$ and memory $O(T(D+H))$, where $L$, $k$, and $H$ are the number of residual blocks, kernel width, and hidden width. The training module ($E_1{+}E_2$) has 7.61M parameters; $E_2$ is discarded, leaving the 3.81M-parameter $E_1$ at inference. At $T=128$, this adds 0.42 ms per ACT decision on an RTX~4090 (6.8 ms for the base policy, 6.2\% overhead) and 2.7 ms on an i7-13700K CPU. GPU latency for $T=32,64,128,256$ is 0.16, 0.27, 0.42, 0.83 ms. We fix $T=128$ in Experiments 2--3; shorter queues occur only during prefix masking.

At inference, an episode-level queue of preceding executed states and actions is encoded at each decision step; future action chunks are never re-encoded. The queue is updated once per executed control step; short histories are left-padded and masked. During joint training, gradients pass through the queued history but not unavailable future actions.

\section{Experiments}
\label{sec:experiments}
We ask whether PMTRM preserves reconstructability, transfers across policy backbones, and improves repetitive manipulation.

\subsection{Experiment 1: Loss Dynamics and Objective Trade-offs}

We examine PMTRM loss dynamics across the three pretraining stages (batch size 64) and subsequent joint policy training. The weighted objective uses $\alpha_{\mathrm{h}}=1.5$, $\beta=0.05$, and $\gamma=8.0$.

\begin{figure*}[h]
    \centering
    \includegraphics[width=0.9\linewidth]{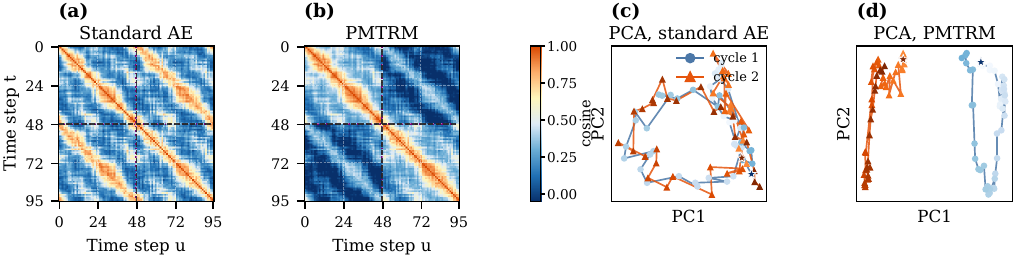}
    \caption{Representation analysis on a phase-aliasing cyclic trajectory. Top: temporal cosine-similarity matrices for a standard autoencoder and PMTRM (the same $[0,1]$ color scale is used; dotted boxes mark cycle-1/cycle-2 aliased-phase blocks measured by Distant Cos. in Table~\ref{tab:pmtrm_ablation}, and dashed lines mark the cycle boundary). Bottom: PCA projections of two cycles; stars mark their starts. PMTRM suppresses off-diagonal similarities while retaining local continuity.}
    \label{fig:temporal_analysis}
\end{figure*}

Figure~\ref{fig:temporal_analysis} visualizes the learned representation on a phase-aliasing cyclic sequence. Unlike the standard autoencoder, PMTRM preserves local continuity while separating distant phases; this qualitative effect is quantified below with a probe on frozen representations.

\begin{table}[t]
\centering
\caption{Ablation of PMTRM design choices on the memory-intensive evaluation (ACT MuJoCo Button Press-and-Release, LIBERO Drawer Open-and-Close, Pour Bottle Twice, and Shell Game). Avg. success follows the metric in Table~\ref{tab:pmtrm_memory}; all variants use the same demonstrations, seeds, and $T=128$.}
\label{tab:pmtrm_ablation}
\resizebox{\columnwidth}{!}{%
\setlength{\tabcolsep}{5pt}
\renewcommand{\arraystretch}{1.08}
\begin{tabular}{@{}lccc@{}}
\toprule
Setting & Avg. Success $\uparrow$ & Recon. MSE $\downarrow$ & Distant Cos. $\downarrow$ \\
\midrule
Single-stage AE + heterogeneity (same budget) & 63.4 & 2.4 & 0.31 \\
GRU + reconstruction (same policy interface) & 64.6 & 3.1 & 0.29 \\
Contrastive temporal loss (same budget) & 67.2 & 3.0 & 0.25 \\
PMTRM w/o synthetic pretraining & 67.7 & 2.8 & 0.23 \\
PMTRM, ACT MuJoCo Markov reach & 91.6 & 1.9 & 0.30 \\
PMTRM, $\beta=0$ & 65.1 & 7.8 & 0.18 \\
PMTRM, $\beta=0.01$ & 68.5 & 4.9 & 0.20 \\
PMTRM, $\beta=0.10$ & 69.2 & 2.1 & 0.27 \\
PMTRM, random mask only (no mirror) & 66.7 & 2.9 & 0.23 \\
\rowcolor[HTML]{CCF2F5}
Full PMTRM, $\beta=0.05$ & \textbf{70.9} & 2.5 & 0.21 \\
\bottomrule
\end{tabular}
}
\vspace{-2mm}
\end{table}

Here Distant Cos. is the mean $s_{k,t}$ over valid pairs with $|t-a_k|>\delta$. The probe on frozen representations is a two-layer MLP (hidden width 64, GELU) trained for 20 epochs with AdamW on 80\% of the training windows and evaluated on the remaining 20\%; labels are press/release or open/close intervals from simulator phase timestamps, and completed-pour count from the annotated action-event index. It gives 98.1\% phase accuracy from $\mathbf Z$, versus 71.4\% from $\mathbf X$ and 76.8\% from single-stage AE. On real Drawer Open-and-Close, $\mathbf Z$ predicts open/close at 94.6\% (raw $\mathbf X$: 79.2\%); on Pour Bottle Twice it predicts pour count at 86.0\% (raw $\mathbf X$: 61.5\%).

Table~\ref{tab:pmtrm_ablation} shows that matched-budget AE, GRU, and contrastive controls underperform PMTRM; all controls share corpus, masks, demonstrations, seeds, and $T=128$. Short-attention/diagonal-SSM/Mamba/action-memory controls obtain 68.8/69.4/69.7/69.2\% at 0.61/0.48/0.46/0.58 ms, versus 70.9\% at 0.42 ms for PMTRM. Across Button CSR / Drawer CSR / Pour Finish / Shell Finish, $\beta=0.01,0.05,0.10$ gives 61/73/70/60, 65/79/80/70, and 64/77/80/60\%, respectively, making $0.05$ a robust compromise. A symmetric negative-cosine penalty reduced local-continuity probe accuracy by 2.3 points and increased MSE by 0.6; negative similarities already separate phases, so the one-sided hinge is retained.

On Markov reaching, ACT/PMTRM obtain $91.8\pm1.0/91.6\pm1.1\%$; periodic tapping gives 88.7/88.1\% (88.5\% with $\alpha_h=0.75$), and $\delta=4,8,12$ varies by under 0.6 points. Mirror/mask/mirror+mask yield 89.8/93.7/98.1\% probe accuracy and 0.24/0.23/0.21 Distant Cos. Dropping proprioception, action, or both lowers success by 1.1/1.8/3.0 points. With ten real rollouts, a 70\% estimate has Wilson 95\% CI $[39.7,89.2]\%$; real gains are directional, while multi-seed simulation is higher-confidence. Distant Cos. is diagnostic, not a value to minimize blindly: around 0.2--0.25 separates aliased phases while preserving continuity; Markov tasks can remain strong near 0.30.

\subsection{Experiment 2: Compatibility with Conventional Action Policies}
\label{subsec:e2}

\begin{figure*}[t] 
    \centering
    \includegraphics[width=\textwidth]{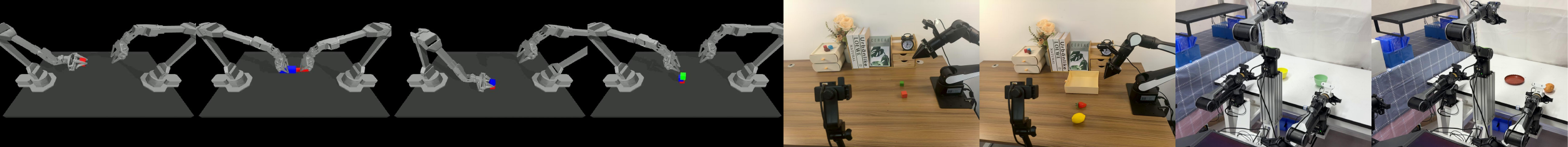}
    \caption{
        Left: The MuJoCo tasks \textit{Transfer Cube}, \textit{Bimanual Insertion}, \textit{Transfer Plus}, and \textit{Stack Two Blocks}. 
        Right: The real-world tasks \textit{Stack Block}, \textit{Store Items}, \textit{Potato Placement}, and \textit{Bowl Stacking}.}
    \label{fig:experiment}
    \vspace{-4mm}
\end{figure*}

\begin{figure}[t] 
    \centering
    \includegraphics[width=0.5\textwidth]{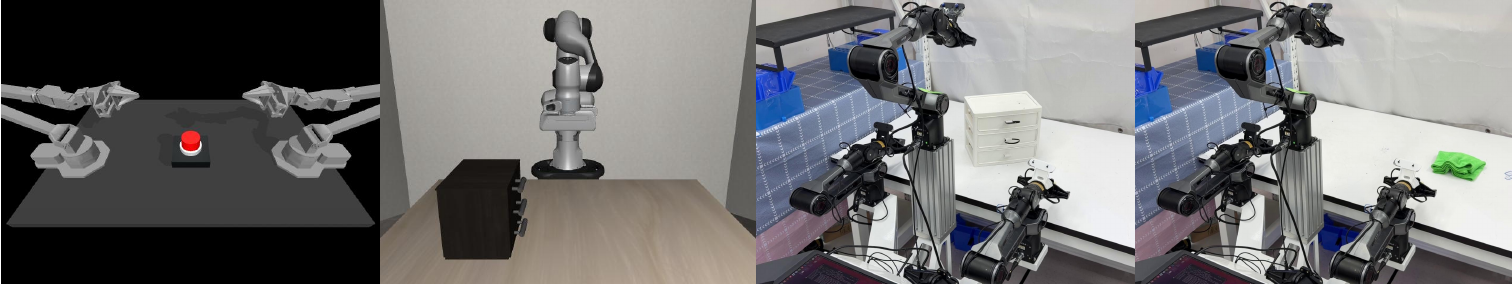}
    \caption{
        Left:
        The simulation tasks \textit{Button Press-and-Release} (MuJoCo) and \textit{Drawer Open-and-Close} (LIBERO).
        Right:
        The real-world tasks
        \textit{Pour Bottle Twice}
        and
        \textit{Shell Game}.
        These tasks require the policy to distinguish repetitive motions and execution phases using historical context rather than only the current observation; the key phase transitions are press versus release, open versus close, first versus second pour, and target tracking after swaps.
        }
    \label{fig:memory_experiment}
    \vspace{-4mm}
\end{figure}

To test generality, we integrate PMTRM with representative policy backbones on the simulated and real-world tasks in Fig.~\ref{fig:experiment}. Touch/Lift/Transfer require end-effector contact, object height $>2$ cm, and placement within 5 cm; Grasp/Contact/Insert require stable grasp, contact within 3 cm, and insertion depth $>90\%$. Finish requires all subtasks within the episode horizon. MuJoCo runs 400 control steps at 20 Hz; real tasks run 300 steps at 10 Hz with reset after a missed grasp or workspace violation. All methods use identical demonstrations and splits, isolating the policy and PMTRM components; action decoders are unchanged.
Simulation uses ten fixed seeds on an RTX~4090; real trials use the same ten randomized poses for every method. Stage~1--3 provide one initialization, then Stage~4 fine-tunes an independent PMTRM copy per backbone.
To isolate modality interactions, removing visual features while retaining PMTRM changes ACT-PMTRM success by $-0.8$ points on simulation, whereas removing proprioceptive/action history while retaining the visual encoder changes it by $-5.4$ points. The module therefore supplies complementary phase information rather than replacing visual grounding.

When PMTRM is integrated with a downstream policy backbone, its auxiliary objective is directly added to the original policy training loss with weights set to $\alpha_{\mathrm{h}}=1.5$, $\beta=0.05$, and $\gamma=8.0$, respectively. 
All policy-specific objectives, schedules, and architectural hyperparameters follow each backbone's default configuration. We retain each backbone's published epoch and batch schedule, select the checkpoint with lowest validation policy loss, and use no test-set early stopping.
Stage~1--3 produce one common initialization; in Stage~4 we clone PMTRM and jointly fine-tune one copy per backbone (no weights or batch statistics are shared across policies). For visual policies, $D_v=256$ for DP and $D_v=512$ for VLA; $P_v$ is learned per backbone and reused across all datasets seen by that policy.

Tables~\ref{tab:pmtrm_sim} and~\ref{tab:pmtrm_real} report the simulation and real-world results. Gains are largest in later-stage \textit{Transfer}, \textit{Insert}, and \textit{Finish} metrics. In Stack Blocks, ACT-PMTRM reduces near-target velocity sign changes from 4.7 to 2.9 per rollout and raises successful grasps from 69\% to 78\%, consistent with fewer approach oscillations. Real-robot experiments use an Airbot Play, a fixed RGB camera, and 7-D Cartesian delta-pose plus gripper actions.

\begin{table*}[t]
\centering
\caption{
Simulation evaluation of PMTRM with different action policy backbones. 
Success rate (\%) $\uparrow$ is reported as mean $\pm$ standard deviation over 10 random seeds; the rightmost subcolumn of each task is end-to-end completion, and all PMTRM variants use $T=128$. 
}
\label{tab:pmtrm_sim}
\resizebox{\textwidth}{!}{%
\setlength{\tabcolsep}{6pt}
\renewcommand{\arraystretch}{1.12}
\begin{tabular}{@{}lccc ccc ccc ccc@{}}
\toprule
\multicolumn{1}{c}{} &
\multicolumn{3}{c}{Transfer Cube (Sim)} &
\multicolumn{3}{c}{Bimanual Insertion (Sim)} &
\multicolumn{3}{c}{Transfer Plus (Sim)} &
\multicolumn{3}{c}{Stack Two Blocks (Sim)} \\
\cmidrule(lr){2-4}\cmidrule(lr){5-7}\cmidrule(lr){8-10}\cmidrule(l){11-13}
 & Touch & Lift & Transfer & Grasp & Contact & Insert &
   Lift & Stack & Finish & Stack & Lift & Finish \\
\midrule

DP (DDPM, CNN)~\cite{chi2023diffusion}
 & \valpm{95.1}{1.2} & \valpm{92.4}{1.3} & \valpm{90.2}{1.9}
 & \valpm{77.4}{2.7} & \valpm{69.0}{3.1} & \valpm{63.2}{3.3}
 & \valpm{62.1}{3.5} & \valpm{52.9}{3.9} & \valpm{52.9}{3.9}
 & \valpm{82.0}{2.7} & \valpm{63.5}{3.2} & \valpm{46.7}{3.6} \\

\rowcolor[HTML]{F0FBFC}
DP-PMTRM
 & \valpm{97.0}{1.0} & \valpm{94.1}{1.1} & \valpm{92.8}{1.5}
 & \valpm{81.0}{2.3} & \valpm{73.6}{2.7} & \valpm{67.8}{3.0}
 & \valpm{66.8}{3.2} & \valpm{58.4}{3.5} & \valpm{58.4}{3.5}
 & \valpm{85.1}{2.4} & \valpm{68.0}{3.0} & \valpm{53.2}{3.4} \\[2pt]

ACT~\cite{zhao2023learning}
 & \valpm{98.4}{0.9} & \valpm{96.2}{1.0} & \valpm{94.8}{2.1}
 & \valpm{81.3}{2.6} & \valpm{73.5}{3.0} & \valpm{68.1}{3.2}
 & \valpm{66.7}{3.6} & \valpm{57.5}{4.0} & \valpm{57.5}{4.0}
 & \valpm{85.8}{2.8} & \valpm{67.3}{3.4} & \valpm{50.4}{3.8} \\

\rowcolor[HTML]{F0FBFC}
ACT-PMTRM
 & \valpm{99.1}{0.5} & \valpm{97.5}{0.8} & \valpm{96.4}{1.5}
 & \valpm{85.7}{2.2} & \valpm{79.4}{2.6} & \valpm{73.8}{2.9}
 & \valpm{72.6}{3.1} & \valpm{63.8}{3.5} & \valpm{63.8}{3.5}
 & \valpm{89.3}{2.3} & \valpm{72.8}{3.0} & \valpm{58.7}{3.4} \\[2pt]

HACT-VQ~\cite{park2024hierarchical}
 & \valpm{98.5}{0.9} & \valpm{97.6}{1.2} & \valpm{96.2}{1.4}
 & \valpm{87.4}{2.4} & \valpm{82.2}{2.7} & \valpm{76.3}{2.8}
 & \valpm{79.2}{2.2} & \valpm{68.5}{3.1} & \valpm{68.5}{3.1}
 & \valpm{90.6}{2.3} & \valpm{76.1}{2.8} & \valpm{55.8}{3.2} \\

\rowcolor[HTML]{F0FBFC}
HACT-VQ-PMTRM
 & \valpm{99.2}{0.5} & \valpm{98.1}{0.8} & \valpm{97.0}{1.0}
 & \valpm{89.0}{1.9} & \valpm{85.9}{2.3} & \valpm{80.1}{2.5}
 & \valpm{81.0}{2.0} & \valpm{71.0}{2.7} & \valpm{71.0}{2.7}
 & \valpm{91.7}{2.0} & \valpm{78.3}{2.5} & \valpm{62.1}{2.9} \\[2pt]

InterACT~\cite{lee2025interact}
 & \valpm{98.2}{0.8} & \valpm{88.4}{1.1} & \valpm{82.1}{2.0}
 & \valpm{88.5}{2.2} & \valpm{78.3}{2.8} & \valpm{44.2}{3.2}
 & \valpm{78.5}{2.4} & \valpm{68.7}{3.4} & \valpm{68.7}{3.4}
 & \valpm{91.9}{2.0} & \valpm{77.1}{2.6} & \valpm{56.9}{2.9} \\

\rowcolor[HTML]{F0FBFC}
InterACT-PMTRM
 & \valpm{98.9}{0.6} & \valpm{91.8}{1.0} & \valpm{86.4}{1.8}
 & \valpm{90.4}{1.9} & \valpm{81.5}{2.4} & \valpm{49.6}{3.0}
 & \valpm{82.1}{2.1} & \valpm{72.2}{3.0} & \valpm{72.2}{3.0}
 & \valpm{93.0}{1.8} & \valpm{80.1}{2.3} & \valpm{61.7}{2.7} \\[2pt]

GR00T N 1.7~\cite{gr00tn1_2025}
 & \valpm{94.5}{1.5} & \valpm{91.2}{2.0} & \valpm{91.6}{2.2}
 & \valpm{83.2}{2.5} & \valpm{73.9}{3.0} & \valpm{69.1}{3.3}
 & \valpm{63.4}{3.1} & \valpm{57.4}{3.9} & \valpm{57.4}{3.9}
 & \valpm{81.3}{2.9} & \valpm{65.4}{3.2} & \valpm{47.5}{3.1} \\

\rowcolor[HTML]{F0FBFC}
GR00T N 1.7-PMTRM
 & \valpm{96.2}{1.1} & \valpm{93.4}{1.6} & \valpm{93.5}{1.8}
 & \valpm{86.0}{2.2} & \valpm{78.1}{2.7} & \valpm{73.6}{3.0}
 & \valpm{68.9}{2.8} & \valpm{62.6}{3.4} & \valpm{62.6}{3.4}
 & \valpm{85.6}{2.5} & \valpm{70.2}{2.9} & \valpm{54.0}{2.8} \\[2pt]



\bottomrule
\end{tabular}
}
\vspace{-3mm}
\end{table*}

\subsection{Experiment 3: Memory and Repetitive Manipulation}
\label{subsec:e3}

\begin{table*}[t]
\centering
\caption{
Real-world evaluation of PMTRM with different action policy backbones. 
Success rates are reported in $[0.0, 1.0]$ with $T=128$ for PMTRM; the rightmost subcolumn of each task is final completion. Final-completion entries use ten Bernoulli trials, while intermediate stage rates pool step outcomes; for the former, a 0.70 estimate has Wilson 95\% CI $[0.397,0.892]$ rather than a standard-deviation interpretation. 
The real-world benchmark includes four physical manipulation tasks: \textit{Stack Block}, \textit{Store Items}, \textit{Potato Placement}, and \textit{Bowl Stacking}.}
\label{tab:pmtrm_real}
\resizebox{\textwidth}{!}{%
\setlength{\tabcolsep}{8pt}
\renewcommand{\arraystretch}{1.12}
\begin{tabular}{@{}lccc ccc ccc ccc@{}}
\toprule
\multicolumn{1}{c}{} &
\multicolumn{3}{c}{Stack Block} &
\multicolumn{3}{c}{Store Items} &
\multicolumn{3}{c}{Potato Placement} &
\multicolumn{3}{c}{Bowl Stacking} \\
\cmidrule(lr){2-4}\cmidrule(lr){5-7}\cmidrule(lr){8-10}\cmidrule(l){11-13}
 & Grasp & Lift & Stack & First & Second & Finish & Grasp & Lift & Place & First & Second & Finish \\
\midrule

DP (DDPM, CNN)~\cite{chi2023diffusion}
 & 0.82 & 0.70 & 0.62
 & 0.79 & 0.61 & 0.54
 & 0.91 & 0.83 & 0.74
 & 0.77 & 0.59 & 0.50 \\

\rowcolor[HTML]{F0FBFC}
DP-PMTRM
 & 0.88 & 0.78 & 0.71
 & 0.86 & 0.86 & 0.64
 & 0.95 & 0.89 & 0.82
 & 0.84 & 0.76 & 0.61 \\[2pt]

ACT~\cite{zhao2023learning}
 & 0.85 & 0.74 & 0.69
 & 0.86 & 0.65 & 0.58
 & 0.93 & 0.86 & 0.80
 & 0.83 & 0.62 & 0.55 \\

\rowcolor[HTML]{F0FBFC}
ACT-PMTRM
 & 0.91 & 0.84 & 0.78
 & 0.91 & 0.78 & 0.70
 & 0.97 & 0.92 & 0.87
 & 0.89 & 0.75 & 0.67 \\[2pt]





GR00T N 1.7~\cite{gr00tn1_2025}
 & 0.86 & 0.76 & 0.70
 & 0.85 & 0.68 & 0.60
 & 0.94 & 0.88 & 0.81
 & 0.83 & 0.65 & 0.57 \\

\rowcolor[HTML]{F0FBFC}
GR00T N 1.7-PMTRM
 & 0.92 & 0.85 & 0.80
 & 0.91 & 0.79 & 0.72
 & 0.98 & 0.94 & 0.89
 & 0.89 & 0.76 & 0.69 \\

\bottomrule
\end{tabular}
}
\vspace{-4mm}
\end{table*}

\begin{table*}[t]
\centering
\caption{
Evaluation on memory-intensive and repetitive manipulation tasks.
For simulation tasks, we report Task Success Rate (TSR, \%) and Cumulative Success Rate (CSR, \%).
For real-world tasks, we report success rate (\%) over 10 randomized closed-loop rollouts; \textit{First Open} and \textit{First Wipe} denote successful completion of the first repeated manipulation, while \textit{Finish} requires completion of the full two-stage sequence.
For reference, a 70\% success rate over 10 Bernoulli trials has a Wilson 95\% confidence interval of approximately $[39.7, 89.2]\%$.
PMTRM uses $T=128$.
}
\label{tab:pmtrm_memory}
\resizebox{\textwidth}{!}{%
\setlength{\tabcolsep}{7pt}
\renewcommand{\arraystretch}{1.12}
\begin{tabular}{@{}lcc cc cc cc@{}}
\toprule
\multicolumn{1}{c}{} &
\multicolumn{2}{c}{Button Press-and-Release (Sim)} &
\multicolumn{2}{c}{Drawer Open-and-Close (Sim)} &
\multicolumn{2}{c}{Open and Close Drawer (Real)} &
\multicolumn{2}{c}{Wipe Table (Real)} \\
\cmidrule(lr){2-3}
\cmidrule(lr){4-5}
\cmidrule(lr){6-7}
\cmidrule(l){8-9}
 & TSR & CSR
 & TSR & CSR
 & First Open & Finish
 & First Wipe & Finish \\
\midrule


HiF-VLA~\cite{lin2026hif}
 & 65.8 & 59.2
 & 68.6 & 60.4
 & 50.0 & 30.0
 & 30.0 & 20.0 \\

MemoryVLA~\cite{shi2025memoryvla}
 & 64.1 & 57.5
 & 72.8 & 64.6
 & 50.0 & 30.0
 & 40.0 & 20.0 \\

MemER~\cite{sridharscaling}
 & 70.6 & 64.4
 & 79.3 & 71.5
 & 60.0 & 40.0
 & 60.0 & 40.0 \\

PrediMem~\cite{lei2026robomemarena}
 & 76.4 & 70.6
 & 84.8 & 78.2
 & 70.0 & 60.0
 & 70.0 & 50.0 \\

PrediMem-PMTRM
 & 79.2 & 73.8
 & 87.4 & 81.6
 & 80.0 & 70.0
 & 80.0 & 60.0 \\


HiF-VLA-PMTRM
 & 69.1 & 62.8
 & 73.9 & 66.7
 & 60.0 & 50.0
 & 70.0 & 50.0 \\

\rowcolor[HTML]{CCF2F5}
\textbf{InterACT-PMTRM}
 & \textbf{85.5} & \textbf{79.7}
 & \textbf{91.8} & \textbf{85.4}
 & \textbf{90.0} & \textbf{80.0}
 & \textbf{80.0} & \textbf{70.0} \\

\bottomrule
\end{tabular}
}
\vspace{-4mm}
\end{table*}

Finally, we evaluate phase-dependent tasks with repeated local observations, where history is required rather than the instantaneous view alone. For each task, training uses the stated demonstrations, validation selects checkpoints and hyperparameters, and test rollouts use held-out scene seeds; no test trajectory is used for tuning.

TSR (task success rate) measures completion of the current task, whereas CSR (cumulative success rate) also requires preceding subtasks to succeed \cite{lei2026robomemarena}. Simulation uses 50 MuJoCo and 100 LIBERO demonstrations; physical tasks use 20 demonstrations and 10 randomized closed-loop rollouts. Demonstrations are split by episode (80/10/10\% train/validation/test); LIBERO drawer variants are split by scene seed, so no trajectory appears in more than one split. Every baseline uses the same demonstrations, seeds, and evaluation rollouts. Published baseline hyperparameters are reused when available; newly attached PMTRM and lightweight controls are tuned on validation only.

As shown in Fig.~\ref{fig:memory_experiment}, we construct four representative memory-intensive tasks, including two simulation tasks and two real-world tasks. The two simulation tasks test canonical state aliasing caused by repeated forward-backward motions, while the two real-world tasks introduce additional phase-progress and spatial-coverage dependencies under physical interaction. These settings require the policy to infer not only the current visual state, but also which stage of the manipulation has already been completed.

\textit{Button Press-and-Release} and LIBERO \textit{Drawer Open-and-Close} contain visually similar observations that correspond to opposite actions at different phases of execution. The real-world \textit{Open and Close Drawer} task further evaluates phase disambiguation under physical interaction: the policy must remember whether the drawer has already been opened in order to determine whether the next action should continue opening or switch to closing. In \textit{Wipe Table}, the robot must retain information about previously wiped regions and recent wiping progress so that subsequent motions cover the remaining area rather than repeatedly acting on the same local region. Together, these tasks evaluate phase disambiguation, progress tracking, and history-dependent execution under repeated observations.

We compare reactive and memory-augmented models, and add a lightweight PrediMem+PMTRM adapter. PMTRM receives executed state/action history while PrediMem retains predictive tokens; CSR improves by 2.4/1.6 points on simulation and final completion by 10 points on each physical task, indicating complementarity.
For PMTRM models, the auxiliary objective is jointly optimized with the original policy loss using the weights in Sec.~III; all backbone settings follow their defaults.

Table~\ref{tab:pmtrm_memory} reports the four memory-intensive tasks. HiF-VLA uses motion tokens, MemoryVLA token memory, MemER keyframe retrieval, and PrediMem predictive memory. MemoryVLA+PMTRM is omitted because its released interface has no aligned executed-action stream and would require tokenizer retraining. InterACT-PMTRM combines local action chunks with phase-indexed history without a keyframe store.

The gains are clearest on repetitive forward-backward and cyclic tasks, where similar local observations can correspond to different actions depending on the preceding trajectory. On Button Press-and-Release, $T=32/64/128/256$ yields 61/70/74/74\% CSR, with 0.16/0.27/0.42/0.83 ms GPU overhead, respectively. A similar saturation trend is observed on the physical tasks: increasing the history window initially improves phase and progress estimation, whereas extending it beyond the relevant interaction horizon provides diminishing returns. Thus $T=128$ provides a favorable latency--accuracy trade-off. A practical rule is to choose $T$ to cover one to two task-phase cycles at the control frequency; increasing it is useful only while CSR or final task success improves faster than latency.

The bounded-history limitation becomes more apparent as the required interaction sequence grows. For \textit{Open and Close Drawer}, errors arise when long or interrupted manipulation sequences make the relevant opening or closing transition fall outside the retained history. For \textit{Wipe Table}, performance degrades when successful completion requires tracking wiping progress over an increasingly long trajectory or maintaining fine-grained coverage information beyond the history window. These failures delimit PMTRM to bounded-history phase and progress disambiguation rather than persistent long-horizon semantic or spatial memory. In these regimes, increasing $T$ beyond 128 gives little benefit once the relevant event lies outside the queue; an external retrieval mechanism, spatial memory, or recurrent full-history module is the appropriate complement.

\section{CONCLUSION}

We proposed PMTRM, a lightweight bounded-history representation module for robotic manipulation. It improves phase disambiguation without retrieval or extra policy tokens, while explicit memories remain preferable for unbounded semantic recall. The module is trained and evaluated here with imitation objectives; in RL, distribution shift and reward-driven phase changes may require annealing $\alpha_h$ or masking heterogeneity when reward-equivalent cycles are detected. Online adaptation is otherwise compatible because the auxiliary losses use executed history only.


\vspace{-3mm}
\bibliographystyle{IEEEtran}
\vspace{3mm}
\bibliography{IEEEexample}
\vspace{3mm}
\end{document}